\documentclass[10pt,twocolumn,letterpaper]{article}

\usepackage{cvpr}              

\usepackage{graphicx}
\usepackage{amsmath}
\usepackage{amssymb}
\usepackage{booktabs}
\usepackage{makecell}
\usepackage{multirow}
\usepackage{xcolor}

\usepackage{comment}
\usepackage[pagebackref,breaklinks,colorlinks]{hyperref}

\usepackage[capitalize]{cleveref}
\crefname{section}{Sec.}{Secs.}
\Crefname{section}{Section}{Sections}
\Crefname{table}{Table}{Tables}
\crefname{table}{Tab.}{Tabs.}

\def\confName{CVPR}
\def\confYear{2022}

\definecolor{Highlight}{HTML}{39b54a}  
\newcommand{\shl}[1]{\footnotesize{#1}}
\newcommand{\hl}[1]{\footnotesize{#1}}

\begin{document}

\title{Swin Transformer V2: Scaling Up Capacity and Resolution}


\author{
Ze Liu\thanks{Equal. $^\dag$Project lead. Ze, Yutong, Zhuliang, Zhenda, Yixuan, Jia are long-term interns at MSRA.}
\quad Han Hu\textsuperscript{*$\dag$} \quad Yutong Lin \quad Zhuliang Yao \quad Zhenda Xie
\quad Yixuan Wei \quad Jia Ning  \\
\quad Yue Cao \quad Zheng Zhang
\quad Li Dong \quad Furu Wei \quad Baining Guo \\
{Microsoft Research Asia}\\
\small{\texttt{\{v-zeliu1,hanhu,t-yutonglin,t-zhuyao,t-zhxie,t-yixuanwei,v-jianing\}@microsoft.com}} \\
\small{\texttt{\{yuecao,zhez,lidong1,fuwei,bainguo\}@microsoft.com}}
}

\maketitle

\begin{abstract}

Large-scale NLP models have been shown to significantly improve the performance on language tasks with no signs of saturation. They also demonstrate amazing few-shot capabilities like that of human beings. This paper aims to explore large-scale models in computer vision. We tackle three major issues in training and application of large vision models, including training instability, resolution gaps between pre-training and fine-tuning, and hunger on labelled data. Three main techniques are proposed: 1) a residual-post-norm method combined with cosine attention to improve training stability; 2) A log-spaced continuous position bias method to effectively transfer models pre-trained using low-resolution images to downstream tasks with high-resolution inputs; 3) A self-supervised pre-training method, SimMIM, to reduce the needs of vast labeled images. Through these techniques, this paper successfully trained a 3 billion-parameter Swin Transformer V2 model, which is the largest dense vision model to date, and makes it capable of training with images of up to 1,536$\times$1,536 resolution. It set new performance records on 4 representative vision tasks, including ImageNet-V2 image classification, COCO object detection, ADE20K semantic segmentation, and Kinetics-400 video action classification. Also note our training is much more efficient than that in Google's billion-level visual models, which consumes 40 times less labelled data and 40 times less training time. Code is available at \url{https://github.com/microsoft/Swin-Transformer}.

\end{abstract}

\section{Introduction}
\label{sec:intro}

Scaling up language models has been incredibly successful. It significantly improves a model’s performance on language tasks~\cite{devlin2018bert,radford2019language,raffel2019t5,Turing-17B,fedus2021switch,Megatron-Turing-530B} and the model demonstrates amazing few-shot capabilities similar to that of human beings~\cite{brown2020language}. Since the BERT large model with 340 million parameters~\cite{devlin2018bert}, language models are quickly scaled up by more than 1,000 times in a few years, reaching 530 billion dense parameters~\cite{Megatron-Turing-530B} and 1.6 trillion sparse parameters~\cite{fedus2021switch}. These large language models are also found to possess increasingly strong few-shot capabilities akin to human intelligence for a broad range of language tasks~\cite{brown2020language}. 

\begin{figure}[t]
    \centering
    \includegraphics[width=1.0\linewidth]{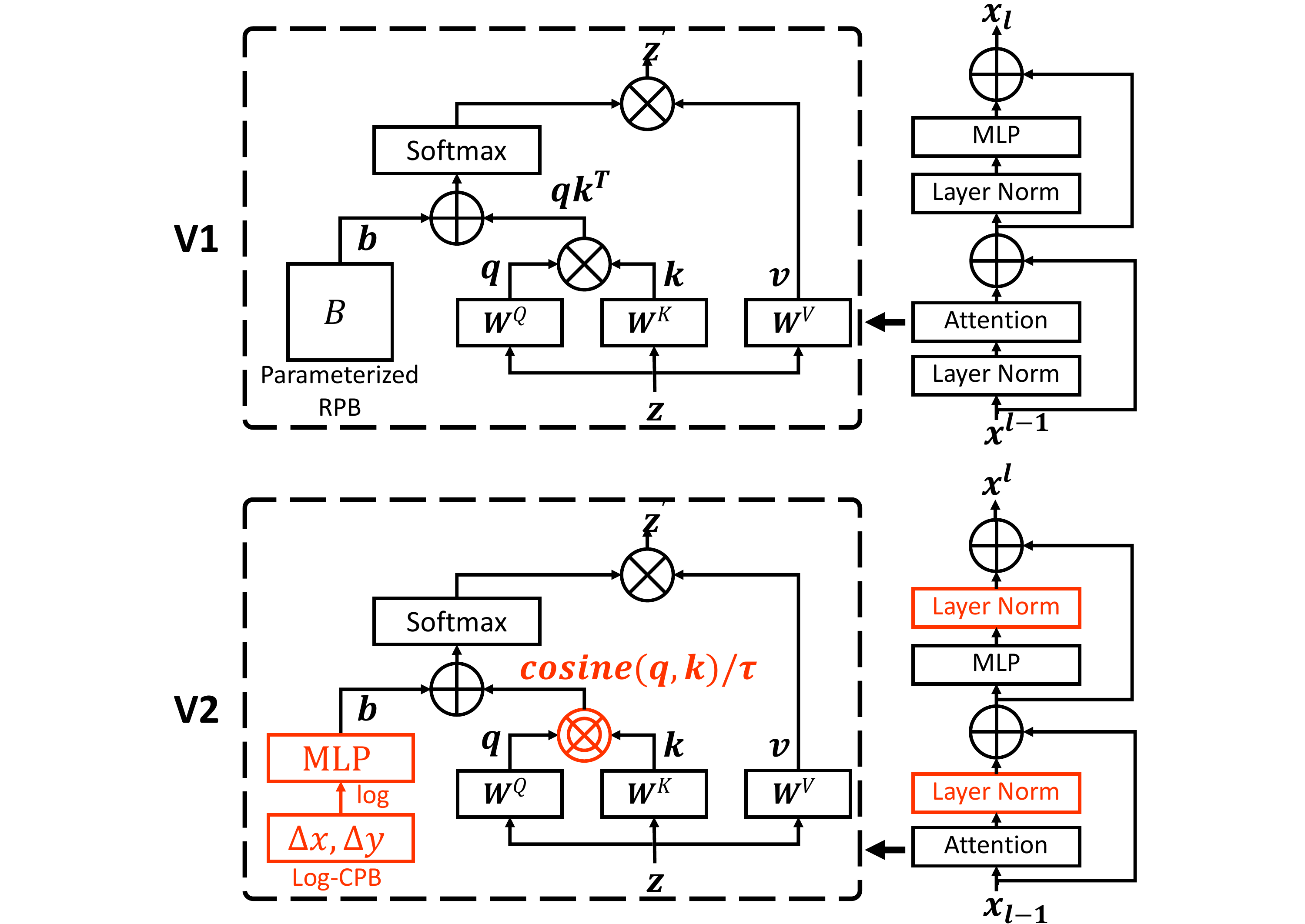}
    \caption{To better scale up model capacity and window resolution, several adaptions are made on the original Swin Transformer architecture (V1): 1) A \emph{res-post-norm} to replace the previous \emph{pre-norm} configuration; 2) A \emph{scaled cosine attention} to replace the original \emph{dot product attention}; 3) A \emph{log-spaced continuous} relative position bias approach to replace the previous \emph{parameterized} approach. Adaptions 1) and 2) make it easier for the model to scale up capacity. Adaption 3) makes the model to be transferred more effectively across window resolutions. The adapted architecture is named Swin Transformer V2.}
    \label{fig:teaser}
    \vspace{-0.5em}
\end{figure}

On the other hand, the scaling up of vision models has been lagging behind. While it has long been recognized that larger vision models usually perform better on vision tasks~\cite{simonyan2014vgg,he2015resnet}, the absolute model size was just able to reach about 1-2 billion parameters very recently~\cite{kolesnikov2019bigtransfer,goyal2021selfsupervised,zhai2021scaling,riquelme2021scaling,dai2021coatnet}. More importantly, unlike large language models, the existing large vision models are applied to the image classification task only~\cite{zhai2021scaling,riquelme2021scaling,dai2021coatnet}. 

To successfully train large and general vision model, we need to address a few key issues. Firstly, our experiments with large vision models reveal an instability issue in training. We find that the discrepancy of activation amplitudes across layers becomes significantly greater in large models. A closer look at the original architecture reveals that this is caused by the output of the residual unit directly added back to the main branch. The result is that the activation values are accumulated layer by layer, and the amplitudes at deeper layers are thus significantly larger than those at early layers. To address this issue, we propose a new normalization configuration, called res-post-norm, which moves the LN layer from the beginning of each residual unit to the backend, as shown in Figure~\ref{fig:teaser}. We find this new configuration produces much milder activation values across the network layers. We also propose a scaled cosine attention to replace the previous dot product attention. The scaled cosine attention makes the computation irrelevant to amplitudes of block inputs, and the attention values are less likely to fall into extremes. In our experiments, the proposed two techniques not only make the training process more stable but also improve the accuracy especially for larger models.

Secondly, many downstream vision tasks such as object detection and semantic segmentation require high resolution input images or large attention windows. The window size variations between low-resolution pre-training and high-resolution fine-tuning can be quite large. The current common practice is to perform a bi-cubic interpolation of the position bias maps~\cite{dosovitskiy2020vit,liu2021swin}. This simple fix is somewhat ad-hoc and the result is usually sub-optimal. We introduce a log-spaced continuous position bias (Log-CPB), which generates bias values for arbitrary coordinate ranges by applying a small meta network on the log-spaced coordinate inputs. Since the meta network takes any coordinates, a pre-trained model will be able to freely transfer across window sizes by sharing weights of the meta network. A critical design of our approach is to transform the coordinates into the log-space so that the extrapolation ratio can be low even when the target window size is significantly larger than that of pre-training. 
The scaling up of model capacity and resolution also leads to prohibitively high GPU memory consumption with existing vision models. To resolve the memory issue, we incorporate several important techniques including zero-optimizer~\cite{rajbhandari2020zero}, activation check pointing~\cite{chen2016training} and a novel implementation of sequential self-attention computation. With these techniques, the GPU memory consumption of large models and resolutions is significantly reduced with only marginal effect on the training speed. 

With the above techniques, we successfully trained a 3 billion Swin Transformer model and effectively transferred it to various vision tasks with image resolution as large as 1,536$\times$1,536, using Nvidia A100-40G GPUs. In our model pre-training, we also employ self-supervised pre-training to reduce the dependency on super-huge labeled data. With 40$\times$ less labelled data than that in previous practice (JFT-3B), the 3 billion model achieves the state-of-the-art accuracy on a broad range of vision benchmarks. Specifically, it obtains 84.0\% top-1 accuracy on the ImageNet-V2 image classification validation set~\cite{recht2019imagenet}, 63.1 / 54.4 box / mask AP on the COCO test-dev set of object detection, 59.9 mIoU on ADE20K semantic segmentation, and 86.8\% top-1 accuracy on Kinetics-400 video action classification, which are +NA\%, +4.4/+3.3, +6.3 and +1.9 higher than the best numbers in the original Swin Transformers~\cite{liu2021swin,liu2021video}, and surpass previous best records by +0.8\% (\cite{zhai2021scaling}), +1.8/+1.4~(\cite{xu2021endtoend}), +1.5 (\cite{bao2021beit}) and +1.4\% (\cite{ryoo2021tokenlearner}).

By scaling up both capacity and resolution of vision models with strong performance on general vision tasks, just like a good language model’s performance on general NLP tasks, we aim to stimulate more research in this direction so that we can eventually close the capacity gap between vision and language models and facilitate the joint modeling of the two domains.

\section{Related Works}

\paragraph{Language networks and scaling up} Transformer has served the standard network since the pioneer work of~\cite{vaswani2017attention}. The exploration of scaling this architecture has since begun, and the progress has been accelerated by the invention of effective self-supervised learning approaches, such as masked or auto-regressive language modeling~\cite{devlin2018bert,radford2019language}, and has been further encouraged by the discovery of a scaling law~\cite{kaplan2020scaling}. Since then, the capacity of language models has increased dramatically by more than 1,000 times in a few years, from BERT-340M to the Megatron-Turing-530B~\cite{raffel2019t5,Turing-17B,brown2020language,Megatron-Turing-530B} and sparse Switch-Transformer-1.6T~\cite{fedus2021switch}. With increased capacity, the accuracy of various language benchmarks has been significantly improved. The zero-shot or few-shot performance is also significantly improved~\cite{brown2020language}, which is a foundation of human generic intelligence.

\paragraph{Vision networks and scaling up} CNNs have long been the standard computer vision networks~\cite{lecun1998lenet, krizhevsky2012alexnet}. Since AlexNet~\cite{krizhevsky2012alexnet}, architectures have become deeper and larger, which has greatly advanced various visual tasks and largely fueled the wave of deep learning in computer vision, such as VGG~\cite{simonyan2014vgg}, GoogleNet~\cite{szegedy2015googlenet} and ResNet~\ cite{he2015resnet}. In the past two years, the CNN architectures have been further scaled up to about 1 billion parameters ~\cite{kolesnikov2019bigtransfer, goyal2021selfsupervised}, however, absolute performance may not be so encouraging, perhaps due to inductive biases in the CNN architecture limiting modeling power. 

Last year, Transformers started taking over one representative visual benchmark after another, including ImageNet-1K image-level classification benchmarks~\cite{dosovitskiy2020vit}, COCO region-level object detection benchmark~\cite{liu2021swin}, ADE20K pixel-level semantic segmentation benchmark~\cite{zheng2020SETR, liu2021swin}, Kinetics-400 video action classification benchmark~\cite{arnab2021vivit}, etc. Since these works, numerous vision Transformer variants have been proposed to improve the accuracy at relatively small scale~\cite{touvron2020deit,li2021localvit,chu2021twins,wang2021pyramid,yuan2021tokenstotoken,zhang2021multiscale,dong2021cswin,yang2021focal,huang2021shuffle,xiao2021early,yuan2021volo}. 
Only a few works have attempted to scale up the vision Transformers~\cite{zhai2021scaling,riquelme2021scaling,dai2021coatnet}. However, they rely on a huge image dataset with classification labels, i.e., JFT-3B, and are only applied to image classification problems.

\paragraph{Transferring across window / kernel resolution} For CNNs, previous works typically fixed kernel size during pre-training and fine-tuning. Global vision Transformers, such as ViT~\cite{dosovitskiy2020vit}, compute attention globally, with the equivalent attention window size linearly proportional to the increased input image resolution. For local vision Transformer architectures, such as Swin Transformer~\cite{liu2021swin}, the window size can be either fixed or changed during fine-tuning. Allowing variable window sizes is more convenient in use, so as to be divisible by the probably variable entire feature map and to tune receptive fields for better accuracy. To handle the variable window sizes between pre-training and fine-tuning, bi-cubic interpolation was the previous common practice~\cite{dosovitskiy2020vit, liu2021swin}. In this paper, we propose a log-spaced continuous position bias approach (Log-CPB) that more smoothly transfers pre-trained model weights at low resolution to deal-with higher resolution windows. 

\paragraph{Study on bias terms} In NLP, the relative position bias method proved beneficial~\cite{raffel2019t5}, compared to the absolute position embedding used in the original Transformer~\cite{vaswani2017attention}. In computer vision, the relative positional bias method is more commonly used~\cite{hu2019localrelation, liu2021swin, yang2021focal}, probably because the spatial relationships of visual signals play a more important role in visual modeling. A common practice is to directly learn the bias values as model weights. There are also a few works particularly study how to set and learn the bias terms~\cite{ke2021rethinking, wu2021rethinking}. 

\paragraph{Continuous convolution and variants} Our Log-CPB approach is also related to earlier works on continuous convolution and variants ~\cite{schutt2017schnet,wang2018continuousconvcvpr,hu2018relation,liu2020closer}, which utilize a meta network to handle irregular data points. Our Log-CPB approach is inspired by these efforts while solving a different problem of transferring relative position biases in vision Transformers across arbitrary window sizes. We also propose log-spaced coordinates to alleviate the difficulty of extrapolation when transferring between large size changes.

\section{Swin Transformer V2}

\subsection{A Brief Review of Swin Transformer}

\label{sec.swin_v1}

Swin Transformer is a general-purpose computer vision backbone that has achieved strong performance in various granular recognition tasks such as region-level object detection, pixel-level semantic segmentation, and image-level image classification. The main idea of Swin Transformer is to introduce several important visual priors into the vanilla Transformer encoder, including hierarchy, locality, and translation invariance, which combines the strength of both: the basic Transformer unit has strong modeling capabilities, and the visual priors make it friendly to a variety of visual tasks. 

\paragraph{Normalization configuration} It is widely known that normalization technologies~\cite{ioffe2015batch, ba2016layer, wu2018group, ulyanov2017instance} are crucial in stably training deeper architectures. The original Swin Transformer inherits the common practice in the language Transformers~\cite{radford2019language} and vanilla ViT~\cite{dosovitskiy2020vit} to utilize a pre-normalization configuration without extensive study, as shown in the figure~\ref{fig:teaser}. In the following subsections, we will examine this default normalization configuration\footnote{There have been a few alternative normalization configurations, such as post-normalization~\cite{vaswani2017attention} and sandwich normalization~\cite{ding2021cogview}. Post-normalization harms training stability~\cite{xiongLN2020}, and sandwich normalization sacrifices representation power due to too many normalization layers.}. 

\paragraph{Relative position bias} is a key component in the original Swin Transformer which introduces an additional parametric bias term to encode the geometric relationship in self-attention calculation:
\begin{equation}
\label{eq.att}
    \text{Attention}(Q, K, V) = \text{SoftMax}(QK^T/\sqrt{d}+B)V,
\end{equation}
where $B \in \mathbb{R}^{M^2 \times M^2}$ is the relative position bias term for each head; $Q, K, V \in \mathbb{R}^{M^2\times d}$ are the \emph{query}, \emph{key} and \emph{value} matrices; $d$ is the \emph{query}/\emph{key} dimension, and $M^2$ is the number of patches in a window. The relative position bias encodes relative spatial configurations of visual elements and is shown critical in a variety of visual tasks, especially for dense recognition tasks such as object detection.

In Swin Transformer, the relative positions along each axis are within the range of $[-M+1, M-1]$ and the relative position bias is parameterized as a bias matrix $\hat{B} \in \mathbb{R}^{(2M-1)\times (2M-1)}$, and the elements in $B$ are taken from $\hat{B}$. When transferring across different window sizes, the learnt relative position bias matrix in pre-training is used to initialize the bias matrix of a different size in fine-tuning by bi-cubic interpolation.

\paragraph{Issues in scaling up model capacity and window resolution} We observe two issues when we scale up the capacity and window resolution of the Swin Transformer.

\begin{table*}[t]
\centering
\small
\addtolength{\tabcolsep}{-3.1pt}
\begin{tabular}{c|c|c|c|c|c|c|c|c|c|c}
\Xhline{1.0pt}
& ImageNet* & \multicolumn{4}{c|}{ImageNet$^\dag$} & \multicolumn{2}{c|}{COCO} & \multicolumn{2}{c}{ADE20k} \\
\cline{2-11}
method & \makecell{W8, I256 \\ top-1 acc} & \makecell{W12, I384 \\ top-1 acc} & \makecell{W16, I512\\top-1 acc} & \makecell{W20, I640 \\top-1 acc} & \makecell{W24, I768\\top-1 acc} & \makecell{W16\\AP$^\text{box}$} & \makecell{W32\\AP$^\text{box}$} & \makecell{W16\\mIoU}& \makecell{W20\\mIoU} &\makecell{W32\\mIoU} \\
\hline
Parameterized position bias~\cite{liu2021swin} & 81.7 & 79.4/82.7 & 77.2/83.0 & 73.2/83.2 & 68.7/83.2 & 50.8 & 50.9 &45.5 & 45.8 & 44.5\\
\hline
Linear-Spaced CPB & \makecell{81.7\\\shl{(+0.0)}} & \makecell{82.0/82.9\\\shl{(+2.6/+0.2)}} & \makecell{81.2/83.3\\\shl{(+4.0/+0.3)}} & \makecell{79.8/83.6\\\shl{(+6.6/+0.4)}} & \makecell{77.6/83.7\\\shl{(+8.9/+0.5)}} & \makecell{50.9\\\shl{(+0.1)}} &\makecell{51.7\\\shl{(+0.8)}}  & \makecell{47.0\\\shl{(+1.5)}}&\makecell{47.4\\\shl{(+1.6)}} & \makecell{47.2\\\shl{(+2.7)}}\\
\hline
Log-Spaced CPB & \makecell{81.8\\\hl{(+0.1)}} &\makecell{82.4/83.2\\\hl{(+3.0/+0.5)}} &\makecell{81.7/83.8\\\hl{(+4.5/+0.8)}} & \makecell{80.4/84.0\\\hl{(+7.2/+0.8)}} & \makecell{79.1/84.2\\\hl{(+10.4/+1.0)}} &\makecell{51.1\\\hl{(+0.3)}} & \makecell{51.8\\\hl{(+0.9)}} & \makecell{47.0\\\hl{(+1.5)}} &\makecell{47.7\\\hl{(+1.9)}} & \makecell{47.8\\\hl{(+3.3)}} \\
\Xhline{1.0pt}
\end{tabular}
\caption{Comparison of different position bias computation approaches using Swin-T. * indicates the top-1 accuracy on ImageNet-1k trained from scratch. The models in * column will be used for testing on the ImageNet-1K image classification task using larger image/window resolutions, marked by $\dag$. For these results, we report both the results w.o./with fine-tuning. These models are also used for fine-tuning on COCO object detection and ADE20K semantic segmentation tasks.}
\label{tab:lcpb}
\vspace{-1em}
\end{table*}

\begin{figure}
    \centering
    \includegraphics[width=0.95\linewidth]{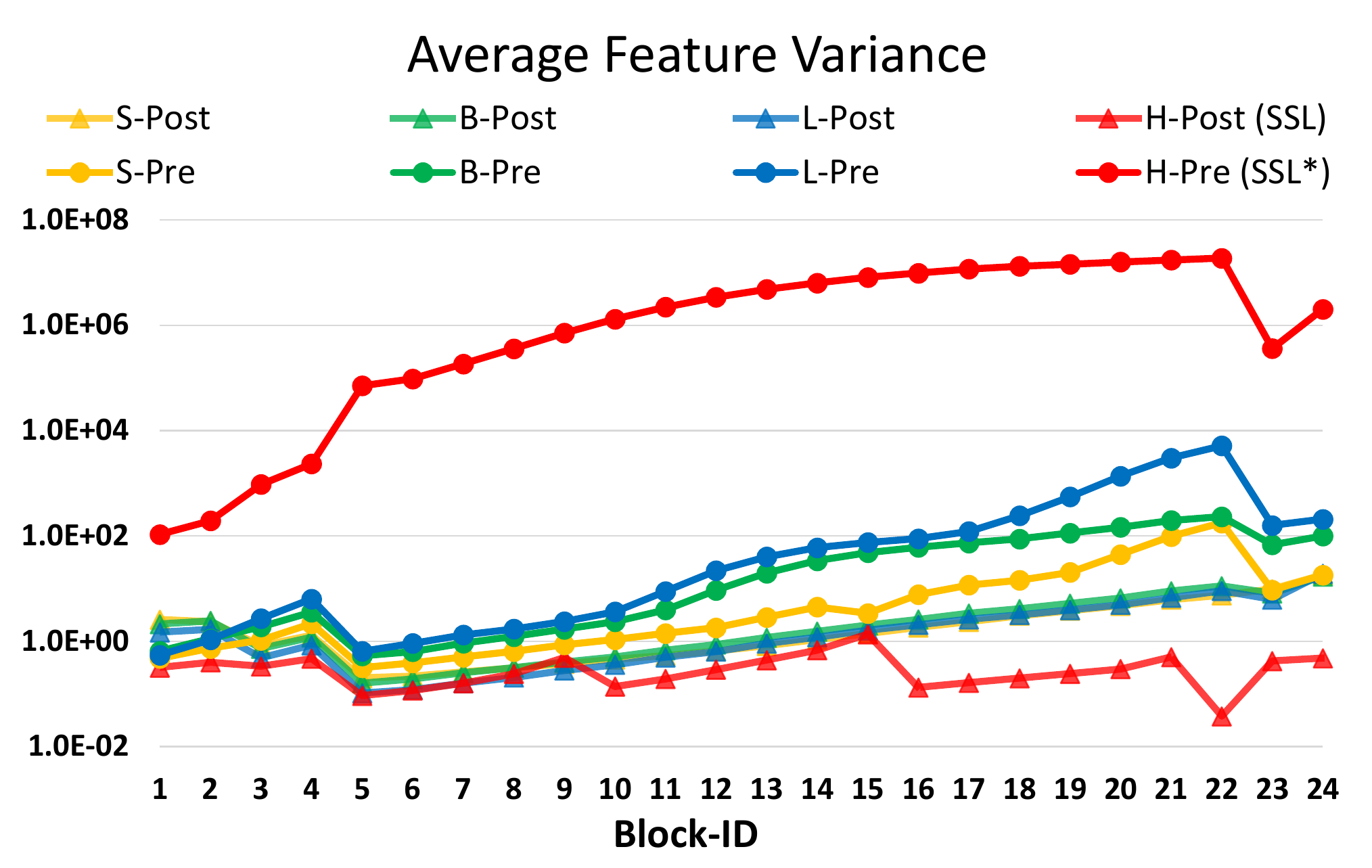}
    \vspace{-1em}
    \caption{The Signal Propagation Plot~\cite{yao2021leveraging,brock2021characterizing} for various model sizes. H-size models are trained at a self-supervised learning phase, and other sizes are trained by an image classification task. * indicates that we use a 40-epoch model before it crashes.}
    \label{fig:act}
    \vspace{-1em}
\end{figure}

\begin{figure}
    \centering
    \includegraphics[width=1.0\linewidth]{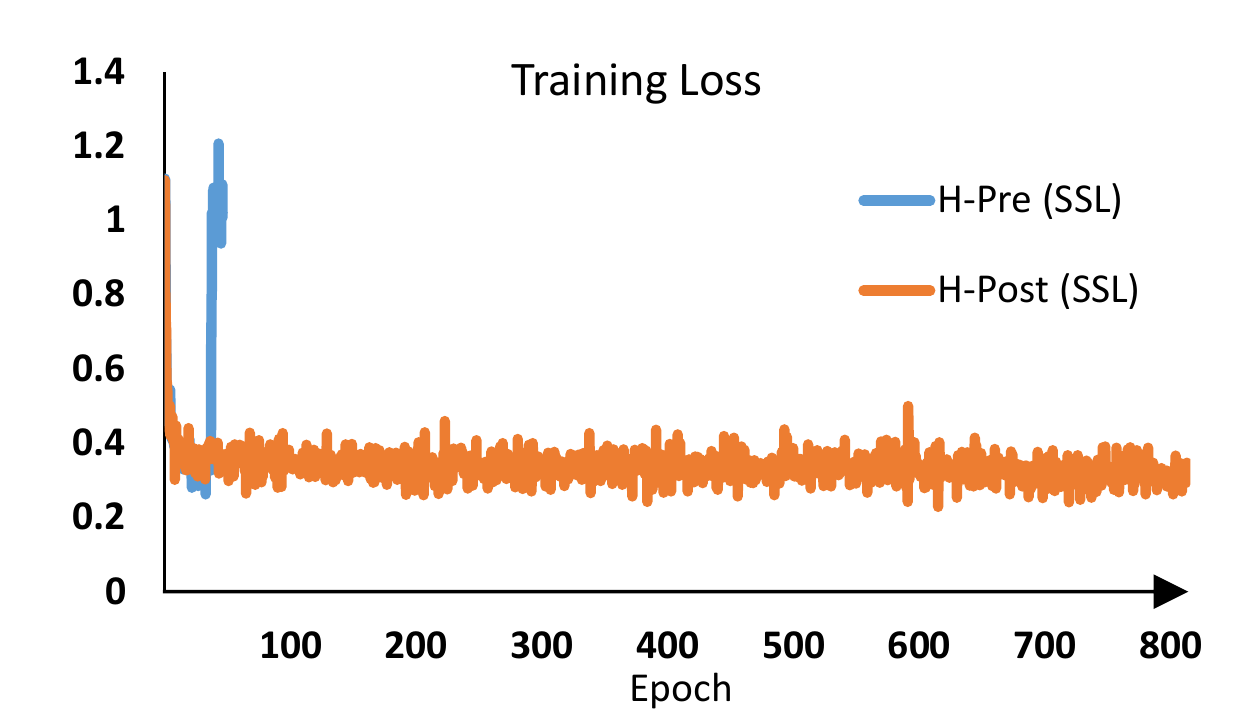}
    \caption{SwinV1-H versus SwinV2-H in training~\cite{simmim}.}
    \label{fig:divergence_of_huge_v1}
    \vspace{-1em}
\end{figure}

\begin{itemize}
\item \emph{An instability issue when scaling up model capacity}. As shown in Figure~\ref{fig:act}, when we scale up the original Swin Transformer model from small size to large size, the activation values at deeper layers increase dramatically. The discrepancy between layers with the highest and the lowest amplitudes has reached an extreme value of $10^4$. When we scale it up further to a huge size (658 million parameters), it cannot complete the training, as shown in Figure~\ref{fig:divergence_of_huge_v1}.

\item \emph{Degraded performance when transferring models across window resolutions}. As shown in the first row of Table~\ref{tab:lcpb}, the accuracy decreases significantly when we directly test the accuracy of a pre-trained ImageNet-1K model ($256\times 256$ images with $8\times 8$ window size) at larger image resolutions and window sizes through the bi-cubic interpolation approach. It may be worth re-examining the relative position bias approach in the original Swin Transformer.

\end{itemize}

In the following subsections, we present techniques to address these issues, including \emph{residual post normalization} and \emph{scaled cosine attention} to address the instability issue, and a \emph{log-spaced continuous position bias} approach to address the issue in transferring across window resolutions.

\subsection{Scaling Up Model Capacity}

As mentioned in Section~\ref{sec.swin_v1}, the original Swin Transformer (and most vision Transformers) adopts a layer norm layer at the beginning of each block, inherited from vanilla ViT. When we scale up the model capacity, a significant increase in activation values is observed at deeper layers. In fact, in a pre-normalization configuration, the output activation values of each residual block are merged directly back to the main branch, and the amplitude of the main branch grows larger and larger at deeper layers. Large amplitude discrepancy in different layers causes training instability.

\paragraph{Post normalization}
To ease this problem, we propose to use a \emph{residual post normalization} approach instead, as shown in Figure~\ref{fig:teaser}. In this approach, the output of each residual block is normalized before merging back into the main branch, and the amplitude of the main branch does not accumulate when the layer goes deeper. As shown in Figure~\ref{fig:act}, the activation amplitudes by this approach are much milder than in the original pre-normalization configuration. 

In our largest model training, we introduce an additional layer normalization layer on the main branch every 6 Transformer blocks, to further stabilize training.

\paragraph{Scaled cosine attention} In the original self-attention computation, the similarity terms of the pixel pairs are computed as a dot product of the \emph{query} and \emph{key} vectors. We find that when this approach is used in large visual models, the learnt attention maps of some blocks and heads are frequently dominated by a few pixel pairs, especially in the \emph{res-post-norm} configuration. To ease this issue, we propose a \emph{scaled cosine attention} approach that computes the attention logit of a pixel pair $i$ and $j$ by a scaled cosine function:
\begin{equation}
\label{eq.att}
    \text{Sim}(\mathbf{q}_i, \mathbf{k}_j) = \text{cos}(\mathbf{q}_i, \mathbf{k}_j) / \tau + B_{ij},
\end{equation}
where $B_{ij}$ is the relative position bias between pixel $i$ and $j$; $\tau$ is a learnable scalar, non-shared across heads and layers. $\tau$ is set larger than 0.01. The cosine function is naturally normalized, and thus can have milder attention values.

\subsection{Scaling Up Window Resolution}

In this subsection, we introduce a log-spaced continuous position bias approach, so that the relative position bias can be smoothly transferred across window resolutions.

\paragraph{Continuous relative position bias} Instead of directly optimizing the parameterized biases, the \emph{continuous} position bias approach adopts a small meta network on the relative coordinates:
\begin{equation}
\label{eq.cpb}
    B (\Delta x, \Delta y) = \mathcal{G} (\Delta x, \Delta y),
\end{equation}
where $\mathcal{G}$ is a small network, e.g., a 2-layer MLP with a ReLU activation in between by default.

The meta network $\mathcal{G}$ generates bias values for arbitrary relative coordinates, and thus can be naturally transferred to fine-tuning tasks with arbitrarily varying window sizes. In inference, the bias values at each relative position can be pre-computed and stored as model parameters, such that the inference is the same as the original parameterized bias approach.

\paragraph{Log-spaced coordinates} 

When transferring across largely varying window sizes, a large portion of the relative coordinate range needs to be extrapolated. To ease this issue, we propose using log-spaced coordinates instead of the original linear-spaced ones:
\begin{equation}
\begin{aligned}
\label{eq.log_coord}
    \widehat{\Delta x} = \text{sign}(x) \cdot \log(1+|\Delta x|), \\
    \widehat{\Delta y} = \text{sign}(y) \cdot \log(1+|\Delta y|),
\end{aligned}
\end{equation}
where $\Delta x$, $\Delta y$ and $\widehat{\Delta x}$, $\widehat{\Delta y}$ are the linear-scaled and log-spaced coordinates, respectively.

By using the log-spaced coordinates, when we transfer the relative position biases across window resolutions, the required extrapolation ratio will be much smaller than that of using the original linear-spaced coordinates. For an example of transferring from a pre-trained $8\times 8$ window size to a fine-tuned $16\times 16$ window size, using the original raw coordinates, the input coordinate range will be from $[-7, 7]\times [-7, 7]$ to $[-15, 15]\times [-15, 15]$. The extrapolation ratio is $\frac{8}{7}=1.14\times$ of the original range. Using log-spaced coordinates, the input range will be from $[-2.079, 2.079]\times [-2.079, 2.079]$ to $[-2.773, 2.773]\times [-2.773, 2.773]$. The extrapolation ratio is $0.33\times$ of the original range, which is an about 4 times smaller extrapolation ratio than that using the original linear-spaced coordinates.

Table~\ref{tab:lcpb} compares the transferring performance of different position bias computation approaches. It can be seen that the log-spaced CPB (continuous position bias) approach performs best, particularly when transferred to larger window sizes.

\subsection{Self-Supervised Pre-training}
Larger models are more data hungry. To address the data hungry problem, previous large vision models typically utilize huge labelled data such as JFT-3B~\cite{zhai2021scaling,riquelme2021scaling,dai2021coatnet}. In this work, we exploit a self-supervised pre-training method, SimMIM~\cite{simmim}, to alleviate the demands on labelled data. By this approach, we successfully trained a powerful Swin Transformer model of 3 billion parameters which achieves state-of-the-art (SOTA) on 4 representative visual benchmarks, by using only 70 million labelled images (1/40 of that in JFT-3B).

\subsection{Implementation to Save GPU Memory}

Another issue lies in the unaffordable GPU memory consumption with a regular implementation when both the capacity and resolution are large. To facility the memory issue, we adopt the following implementations:
\begin{itemize}
    \item \emph{Zero-Redundancy Optimizer (ZeRO)}~\cite{rajbhandari2020zero}. In a general data-parallel implementation of optimizers, the model parameters and optimization states are broadcasted to every GPU. This implementation is very unfriendly on GPU memory consumption, for example, a model of 3 billion parameters will consume 48G GPU memory when an AdamW optimizer and fp32 weights/states are used. With a ZeRO optimizer, the model parameters and the corresponding optimization states will be split and distributed to multiple GPUs, which significantly reduces memory consumption. We adopt the DeepSpeed framework and use the ZeRO stage-1 option in our experiments. This optimization has little effect on training speed.
    \item \emph{Activation check-pointing}~\cite{chen2016training}. Feature maps in the Transformer layers also consume a lot of GPU memory, which can create bottlenecks when image and window resolutions are high. The activation check-pointing technology can significantly reduce the memory consumption, while the training speed is up to 30\% slower.
    \item \emph{Sequential self-attention computation}. To train large models on very large resolutions, for example, an image of 1,536$\times$1,536 resolution with a window size of 32$\times$32, regular A100 GPUs (40GB memory) are still unaffordable, even with the above two optimization technologies. We found that in this case, the self-attention module constitutes a bottleneck. To alleviate this problem, we implement self-attention computation sequentially, instead of using the previous batch computation approach. This optimization is applied to the layers in the first two stages and has little impact on the overall training speed.
\end{itemize}

With these implementations, we managed to train a 3B model using the Nvidia A100-40G GPUs for COCO object detection with an input image resolution of 1,536$\times$1,536, and Kinetics-400 action classification with an input resolution of $320\times 320 \times 8$.

\subsection{Model configurations}

We maintain the stage, block, and channel settings of the original Swin Transformer for 4 configurations of Swin Transformer V2:
\begin{itemize}
    \item SwinV2-T: $C$ = $96$, \#. block = $\{2, 2, 6, 2\}$
    \item SwinV2-S/B/L: $C$=$96/128/192$, \#.block=$\{2, 2, 18, 2\}$
\end{itemize}
with $C$ the number of channels in the first stage. 

We further scale up Swin Transformer V2 to its huge size and giant size, with 658 million parameters and 3 billion parameters, respectively:
\begin{itemize}
    \item SwinV2-H: $C=352$, \#. block = $\{2, 2, 18, 2\}$
    \item SwinV2-G: $C=512$, \#. block = $\{2, 2, 42, 4\}$
\end{itemize}
For SwinV2-H and SwinV2-G, we add an additional layer normalization layer on the main branch every 6 layers. To save experimental time, we only employ SwinV2-G for large-scale experiments. 
SwinV2-H is employed for another parallel study about self-supervised learning~\cite{simmim}.

\section{Experiments}

\subsection{Tasks and Datasets}

We conduct experiments on ImageNet-1K image classification (V1 and V2)~\cite{deng2009imagenet,recht2019imagenet}, COCO object detection~\cite{lin2014coco}, and ADE20K semantic segmentation~\cite{zhou2018semantic}. For the 3B model experiments, we also report the accuracy on Kinetics-400 video action recognition~\cite{kay2017kinetics}.

\begin{itemize}
    \item \emph{Image classification}. ImageNet-1K V1 and V2 val are used~\cite{deng2009imagenet,recht2019imagenet} for evaluation. ImageNet-22K~\cite{deng2009imagenet} which has 14M images and 22K categories is optionally employed for pre-training.  For the pre-training our largest model SwinV2-G, a privately collected ImageNet-22K-ext dataset with 70 million images is used. For this dataset, a duplicate removal process~\cite{radford2021clip} is conducted to exclude overlapping images with ImageNet-1K V1 and V2 validation sets.
    \item \emph{Object detection}. COCO~\cite{lin2014coco} is used for evaluation. For our largest model experiments, we employ an additional detection pre-training phase using Object 365 v2 dataset~\cite{Shao_2019_ICCV}, in-between the image classification pre-training phase and the COCO fine-tuning phase.
    \item \emph{Semantic segmentation}. ADE20K~\cite{zhou2018semantic} is used.
    \item \emph{Video action classification}. Kinetics-400 (K400)~\cite{kay2017kinetics} is used in evaluation. 
\end{itemize}

The pre-training and fine-tuning settings will be detailed in Appendix.

\begin{table*}[t]
    \centering
    \small
    \addtolength{\tabcolsep}{-2.0pt}
    \begin{tabular}{ccccccccc}
    \Xhline{1.0pt}
    Method & param & \makecell{ pre-train\\images} & \makecell{ pre-train\\length (\#im)} & \makecell{pre-train\\im size} & \makecell{pre-train\\time} & \makecell{fine-tune\\im size} & \makecell{ImageNet-1K-V1\\top-1 acc} & \makecell{ImaegNet-1K-V2\\top-1 acc}\\
    \hline
    SwinV1-B & 88M & IN-22K-14M & 1.3B & 224$^2$ & $<$30$^\dag$ & 384$^2$ &86.4 &  76.58\\
    SwinV1-L & 197M & IN-22K-14M & 1.3B & 224$^2$ & $<$10$^\dag$ & 384$^2$ &  87.3 & 77.46\\
    \hline
    ViT-G~\cite{zhai2021scaling} & 1.8B & JFT-3B & 164B & 224$^2$ & $>$30k & 518$^2$ & 90.45 & 83.33 \\
    V-MoE~\cite{riquelme2021scaling} & 14.7B* & JFT-3B & - & 224$^2$ & 16.8k & 518$^2$ & 90.35 & - \\
    CoAtNet-7~\cite{dai2021coatnet} & 2.44B & JFT-3B & - & 224$^2$ & 20.1k & 512$^2$ & \textbf{90.88} & - \\
    \hline
    SwinV2-B & 88M & IN-22K-14M & 1.3B & 192$^2$ & $<$30$^\dag$ & 384$^2$ & 87.1 & 78.08\\
    SwinV2-L & 197M & IN-22K-14M & 1.3B & 192$^2$ & $<$20$^\dag$ & 384$^2$ & 87.7 & 78.31\\
    { SwinV2-G} & 3.0B & IN-22K-ext-70M & 3.5B & 192$^2$ & $<$0.5k$^\dag$ & 640$^2$  & 90.17 & \textbf{84.00} \\
    \Xhline{1.0pt}
    \end{tabular}
    \caption{Comparison with previous largest vision models on ImageNet-1K V1 and V2 classification. * indicates the sparse model; the ``pre-train time'' column is measured by the TPUv3 core days with numbers copied from the original papers. $\dag$ That of SwinV2-G is estimated according to training iterations and FLOPs.}
    \label{tab:sota_imagenet}
    \vspace{-1em}
\end{table*}

\begin{table}[t]
    \centering
    \small
    \addtolength{\tabcolsep}{-4.0pt}
    \begin{tabular}{c|cc|cc|cc}
    \Xhline{1.0pt}
     \multirow{2}{*}{Method} & \multirow{2}{*}{\makecell{train\\I(W) size}} & \multirow{2}{*}{\makecell{test\\I(W) size}} &\multicolumn{2}{c|}{{\scriptsize mini-val (AP)}} & \multicolumn{2}{c}{{\scriptsize test-dev (AP)}}\\
    & & & box &  mask & box &  mask\\
    \hline
    CopyPaste\cite{ghiasi2020copy} & 1280(-) & 1280(-) & 57.0 & 48.9 & 57.3 & 49.1 \\
    SwinV1-L\cite{liu2021swin} & 800(7)& ms(7)& 58.0 & 50.4 & 58.7 & 51.1\\
    YOLOR\cite{wang2021learn} & 1280(-) & 1280(-) & -&- & 57.3 & -\\
    CBNet\cite{liang2021cbnetv2} & 1400(7) & ms(7) & 59.6& 51.8& 60.1 & 52.3\\
    DyHead\cite{dai2021dynamic} & 1200(-) & ms(-)& 60.3 & - & 60.6 & - \\
    {\scriptsize SoftTeacher}\cite{xu2021endtoend} & 1280(12) & ms(12) & 60.7 & 52.5 & 61.3 & 53.0 \\
    \hline
    \multirow{3}{*}{\makecell{SwinV2-L\\(HTC++)}} & \multirow{3}{*}{1536(32)}& 1100(32) & 58.8 & 51.1& - & -\\
     & & 1100 (48) & 58.9 &  51.2& - & -\\
     & & ms (48) & 60.2 & 52.1 &60.8 &52.7\\
    \hline
    \multirow{3}{*}{\makecell{SwinV2-G\\(HTC++)}} & \multirow{3}{*}{1536(32)}& 1100(32) & 61.7 & 53.3 & - & -\\
     & & 1100 (48) & 61.9 & 53.4 & - & -\\
    & & ms (48) & \textbf{62.5} & \textbf{53.7} & \textbf{63.1} & \textbf{54.4}\\
    \Xhline{1.0pt}
    \end{tabular}
    \caption{Comparison with previous best results on COCO object detection and instance segmentation. I(W) indicates the image and window size. ms indicate multi-scale testing is employed.}
    \label{tab:sota_coco}
    \vspace{-1em}
\end{table}

\begin{table}[t]
    \centering
    \small
    \addtolength{\tabcolsep}{-4.0pt}
    \begin{tabular}{c|cc|c}
    \Xhline{1.0pt}
     Method & {\makecell{train I(W) size}} & {\makecell{test I(W) size}} & mIoU \\
    \hline
    SwinV1-L\cite{liu2021swin} & 640(7) & 640(7) & 53.5*\\
    Focal-L\cite{yang2021focal} & 640(40) & 640(40) & 55.4*\\
    CSwin-L\cite{dong2021cswin} & 640(40) & 640(40) & 55.7*\\
    MaskFormer\cite{cheng2021maskformer} & 640(7) & 640(7) & 55.6*\\
    FaPN\cite{huang2021fapn} & 640(7) & 640(7) & 56.7* \\
    BEiT\cite{bao2021beit} & 640(40) & 640(40) & 58.4* \\
    \hline
    \makecell{SwinV2-L\\(UperNet)} & 640(40) & 640(40) & 55.9* \\
    \hline
    \multirow{3}{*}{\makecell{SwinV2-G\\(UperNet)}} & \multirow{3}{*}{640(40)}& 640(40) & 59.1 \\
     & & 896 (56) & 59.3 \\
    & & 896 (56) & \textbf{59.9}*\\
    \Xhline{1.0pt}
    \end{tabular}
    \caption{Comparison with previous best results on ADE20K semantic segmentation. * indicates multi-scale testing is used.}
    \label{tab:sota_ade}
    \vspace{-1em}
\end{table}

\begin{table}[htb]
    \centering
    \small
    \addtolength{\tabcolsep}{-4.0pt}
    \begin{tabular}{c|cccc}
    \Xhline{1.0pt}
     Method & {\makecell{train I(W) size}} & {\makecell{test I(W) size}} & views & top-1\\
    \hline
    ViViT\cite{arnab2021vivit} & \makecell{-(-)} & -(-) & 4$\times$3 & 84.8 \\
    SwinV1-L\cite{liu2021video} & \makecell{480(12)$^2\times$16(8)} & \makecell{480(12)$^2\times$16(8)} & 10$\times$5 & 84.9 \\
    TokenLearner\cite{ryoo2021tokenlearner} & \makecell{256(8)$^2\times$64(64)} & \makecell{256(8)$^2\times$64(64)} & 4$\times$3 & 85.4\\
    \hline
    \multirow{3}{*}{\makecell{Video-SwinV2-G}} & \multirow{3}{*}{\makecell{320(20)$^2\times$8(8)}} & \makecell{320(20)$^2\times$8(8)} & 1$\times$1 & 83.2\\
     & & \makecell{384(24)$^2\times$8(8)} & 1$\times$1 & 83.4 \\
     & & \makecell{384(24)$^2\times$8(8)} & 4$\times$5 & \textbf{86.8} \\
    \Xhline{1.0pt}
    \end{tabular}
    \caption{Comparison with previous best results on Kinetics-400 video action classification.}
    \label{tab:sota_kinetics}
    \vspace{-1em}
\end{table}

\subsection{Scaling Up Experiments}

We first present the results on various representative visual benchmarks by scaling up models to 3 billion parameters and to high image/window resolutions.

\paragraph{Settings for SwinV2-G experiments}

We adopt a smaller $192\times 192$ image resolution in pre-training to save on training costs. We take a 2-step pre-training approach. First, the model is pre-trained using a self-supervised method~\cite{simmim} on the ImageNet-22K-ext dataset by 20 epochs. Second, the model is further pre-trained by 30 epochs using the image classification task on this dataset. Detailed pre-training and fine-tuning setups are described in the appendix.

In the following paragraphs, we report the accuracy of SwinV2-G on representative vision benchmarks. Note that since our main goal is to explore how to feasibly scale up model capacity and window resolution, and whether the vision tasks can benefit from significantly larger capacity, we did not particularly align complexities or pre-training data in comparisons.

\paragraph{ImageNet-1K image classification results}

Table~\ref{tab:sota_imagenet} compares the SwinV2-G model with previously largest/best vision models on ImageNet-1K V1 and V2 classification. SwinV2-G is the largest dense vision model to present. It achieves a top-1 accuracy of 84.0\% on the ImageNet V2 benchmark, which is +0.7\% higher than previous best one (83.3\%). Our accuracy on ImageNet-1K V1 is marginally lower (90.17\% vs 90.88\%). The performance difference might come from different degrees of dataset over-tuning~\cite{recht2019imagenet}. Also note we employ much less training iterations and lower image resolutions than those in previous efforts, while performing very well.

We also compare the SwinV2-B and SwinV2-L to the original SwinV1-B and SwinV1-L, respectively, where a +0.8\% and +0.4\% gains are observed. The shrunken gains by SwinV2-L than that of SwinV2-B may imply that if exceeding this size, more labeled data, stronger regularization, or advanced self-supervised learning methods are required.

\vspace{-0.5em}
\paragraph{COCO object detection results}

Table~\ref{tab:sota_coco} compares the SwinV2-G model with previous best results on COCO object detection and instance segmentation. It achieves 63.1/54.4 box/max AP on COCO test-dev, which is +1.8/1.4 higher than previous best numberw (61.3/53.0 by \cite{xu2021endtoend}). This suggests that scaling up vision model is beneficial for the dense vision recognition task of object detection. Our approach can use a different window size at test to additionally benefit, probably attributed to the effective Log-spaced CPB approach.

\vspace{-0.5em}

\paragraph{ADE20K semantic segmentation results}

Table~\ref{tab:sota_ade} compares the SwinV2-G model with previous best results on the ADE20K semantic segmentation benchmark. It achieves 59.9 mIoU on ADE20K val set, +1.5 higher than the previous best number (58.4 by \cite{bao2021beit}). This suggests scaling up vision model is beneficial for pixel-level vision recognition tasks. Using a larger window size at test time can additionally bring +0.2 gains, probably attributed to the effective Log-spaced CPB approach.

\vspace{-0.5em}

\paragraph{Kinetics-400 video action classification results}

Table~\ref{tab:sota_kinetics} compares the SwinV2-G model with previous best results on the Kinetics-400 action classification benchmark. It achieves 86.8\% top-1 accuracy, +1.4\% higher than previous best number~\cite{ryoo2021tokenlearner}. This suggests that scaling up vision models also benefits video recognition tasks. In this scenario, using a larger window size at test time can also bring additional benefits of +0.2\%, probably attributed to the effective Log-spaced CPB approach.

\begin{table}[t]
    \centering
    \small
    \begin{tabular}{c|c|c|c}
    \Xhline{1.0pt}
    Backbone & res-post-norm & \makecell{scaled cosine\\attention} & \makecell{ImageNet \\ top-1 acc} \\
    \hline
    \multirow{3}{*}{Swin-T} & & & 81.5 \\
    & $\checkmark$ & & 81.6\\
    & $\checkmark$ & $\checkmark$ & \textbf{81.7} \\
    \hline
    \multirow{3}{*}{Swin-S} & & &83.2 \\
    & $\checkmark$ & & 83.3\\
    & $\checkmark$ & $\checkmark$ & \textbf{83.6} \\
    \hline
    \multirow{3}{*}{Swin-B} & & &83.6 \\
    & $\checkmark$ & & 83.8\\
    & $\checkmark$ & $\checkmark$ & \textbf{84.1} \\  
     \hline
     \hline
    \multirow{2}{*}{ViT-B} & & &82.2 \\
    & $\checkmark$ & $\checkmark$ & \textbf{82.6} \\ 
    \Xhline{1.0pt}
    \end{tabular}
    \caption{Ablation on res-post-norm and cosine attention.}
    \label{tab:postnorm_cosatt}
    \vspace{-1em}
\end{table}

\begin{table}[t]
    \centering
    \small
    \addtolength{\tabcolsep}{-2.0pt}
    \begin{tabular}{c|c|c|c|c}
    \Xhline{1.0pt}
    Backbone & pre-norm & \makecell{sandwich}~\cite{ding2021cogview}  & \makecell{post-norm}~\cite{vaswani2017attention} & \makecell{our}  \\
    \hline
    Swin-S & 83.2 & 82.6 & 83.3 & \textbf{83.6} \\
    Swin-B & 83.6 & - & 83.6 & \textbf{84.1} \\
    \Xhline{1.0pt}
    \end{tabular}
    \caption{Comparison with other normalization methods. The post-norm method diverges at the default learning rate, and we use 1/4 of the default learning rate for this method. Sandwich performs worse than ours, probably because it sacrifices expressiveness.}
    \label{tab:norm}
    \vspace{-1em}
\end{table}

\subsection{Ablation Study}

\paragraph{Ablation on res-post-norm and scaled cosine attention} Table~\ref{tab:postnorm_cosatt} ablates the performance of applying the proposed res-post-norm and scaled cosine attention approaches to Swin Transformer. Both techniques improve the accuracy at all the tiny, small and base size, and the overall improvements are +0.2\%, +0.4\% and +0.5\% respectively, indicating the techniques are more beneficial for larger models. It also turns out to benefit ViT architecture (+0.4\%). The proposed normalization approach also performs better than some other normalization methods, as shown in Table~\ref{tab:norm}.

More importantly, the combination of post-norm and scaled cosine attention stabilize the training. As shown in Figure~\ref{fig:act}, while the activation values at deeper layers for the original Swin Transformer are almost exploded at large (L) size, those of the new version have much milder behavior. On a huge size model, the self-supervised pre-training~\cite{simmim} diverges using the original Swin Transformer, while it trains well by a Swin Transformer V2 model.

\vspace{-0.5em}

\paragraph{Scaling up window resolution by different approaches} Table~\ref{tab:lcpb} and \ref{tab:ablate_cpb_size} ablate the performance of 3 approaches by scaling window resolutions from $256\times 256$ in pre-training to larger sizes in 3 down-stream vision tasks of ImageNet-1K image classification, COCO object detection, and ADE20K semantic segmentation, respectively. It can be seen that: 1) Different approaches have similar accuracy in pre-training (81.7\%-81.8\%); 2) When transferred to down-stream tasks, the two continuous position bias (CPB) approaches perform consistently better than the parameterized position bias approach used in Swin Transformer V1. Compared to the linear-spaced approach, the log-spaced version is marginally better; 3) The larger the change in resolutions between pre-training and fine-tuning, the larger the benefit of the proposed log-spaced CPB approach. 

In Table~\ref{tab:lcpb} and \ref{tab:ablate_cpb_size}, we also report the accuracy using targeted window resolutions without fine-tuning (see the first number in each column in the ImageNet-1K experiments). The recognition accuracy remains not bad even when the window size is enlarged from $8$ to $24$ (78.9\% versus 81.8\%), while the top-1 accuracy of the original approach significantly degrades from 81.7\% to 68.7\%. Also note that without fine-tuning, using a window size of $12$ that the pre-trained model has never seen before can even be +0.4\% higher that the original accuracy. This suggests that we can improve accuracy through test-time window adjustment, as also observed in Table~\ref{tab:sota_coco}, \ref{tab:sota_ade} and \ref{tab:sota_kinetics}.

\begin{table}[t]
    \centering
    \small
\addtolength{\tabcolsep}{-2.5pt}
    \begin{tabular}{c|c|c|c|c}
    \Xhline{1.0pt}
    & & ImageNet* & \multicolumn{2}{c}{ImageNet$^\dag$} \\
    \cline{3-5}
    Backbone & L-CPB & \makecell{W8, I256} & \makecell{W12, I384 } & \makecell{W16, I512}\\
    \hline
    \multirow{2}{*}{SwinV2-S} &   & 83.7 & 81.8/84.5 & 79.4/84.9\\
    & $\checkmark$ & 83.7 & 84.1/84.8 & 82.9/85.4 \\
    \hline
    \multirow{2}{*}{SwinV2-B} &  & 84.1 & 82.9/85.0 & 81.0/85.3\\
    & $\checkmark$ & 84.2 & 84.5/85.1 & 83.8/85.6 \\
    \Xhline{1.0pt}
    \end{tabular}
    \caption{Ablation on Log-CPB using different model sizes.}
    \label{tab:ablate_cpb_size}
    \vspace{-2em}
\end{table}

\section{Conclusion}

We have presented techniques for scaling Swin Transformer up to 3 billion parameters and making it capable of training with images of up to 1,536$\times$1,536 resolution, including the \emph{res-post-norm} and \emph{scaled cosine attention} to make the model easier to be scaled up in capacity, as well a log-spaced continuous relative position bias approach which lets the model more effectively transferred across window resolutions. The adapted architecture is named Swin Transformer V2, and by scaling up capacity and resolution, it sets new records on 4 representative vision benchmarks. By these strong results, we hope to stimulate more research in this direction so that we can eventually close the capacity gap between vision and language models and facilitate the joint modeling of the two domains.

\section*{Acknowledgement}

We thank many colleagues at Microsoft for their help, in particular, Eric Chang, Lidong Zhou, Jing Tao, Aaron Zhang, Edward Cui, Bin Xiao, Lu Yuan, Peng Cheng, Fan Yang for useful discussion and the help on GPU resources and datasets.

\renewcommand{\thesection}{A\arabic{section}}
\setcounter{section}{0}

\section{Experimental Settings for Ablation}

This section describes the experimental settings for ablation, including models of SwinV2-T, SwinV2-S, and SwinV2-B, and tasks of ImageNet-1K image classification, COCO object detection and ADE semantic segmentation. 

\subsection{ImageNet-1K Pre-training}

All ablation study use the ImageNet-1K image classification task for pre-training. We adopt an input image size (window size) of 256$\times$256 (8$\times$8)\footnote{Most of our experiments have the window size as an even number to make the window shifting offset divisible by the window size. Nevertheless, an odd number of window size also works well, as is right the case in the original Swin Transformer ($7\times 7$).}. Following \cite{liu2021swin}, we employ an AdamW~\cite{loshchilov2017decoupled} optimizer for 300 epochs using a cosine decay learning rate scheduler with 20 epochs of linear warm-up. A batch size of 1024, an initial learning rate of $1\times10^{-3}$, a weight decay of 0.05, and gradient clipping with a max norm of 5.0 are used. Augmentation and regularization strategies include RandAugment~\cite{cubuk2020randaugment}, Mixup~\cite{zhang2017mixup}, Cutmix~\cite{yun2019cutmix}, random erasing~\cite{zhong2020random} and stochastic depth~\cite{huang2016deep}. An increasing degree of stochastic depth augmentation is employed for larger models, i.e. $0.2, 0.3, 0.5$ for tiny, small, and base models, respectively. 

\subsection{Fine-tuning on various tasks}

\paragraph{ImageNet-1K image classification} For ImageNet-1K image classification experiments, we conduct a fine-tuning step if the input image resolution is larger than that in the pre-training step. The fine-tuning lasts for 30 epochs, with an AdamW~\cite{loshchilov2017decoupled} optimizer, a cosine decay learning rate scheduler with an initial learning rate of $4\times10^{-5}$, a weight decay of $1\times10^{-8}$, and the same data augmentation and regularizations as those in the first stage.

\paragraph{COCO object detection}
We use cascade mask R-CNN~\cite{he2017mask,cai2018cascade} implemented in mmdetection~\cite{chen2019mmdetection} as the object detection framework. In training, a multi-scale augmentation~\cite{carion2020detr,sun2020sparsercnn} with the shorter side between 480 and 800 and the longer side of 1333 is used. The window size is set 16$\times$16. An AdamW~\cite{loshchilov2017decoupled} optimizer with an initial learning rate of $1\times10^{-4}$, a weight decay of 0.05, a batch size of 16, and a 3$\times$ scheduler are used.

\paragraph{ADE20K semantic segmentation}
We adopt an image size (window size) of 512$\times$512 (16$\times$16). In training, we employ an AdamW~\cite{loshchilov2017decoupled} optimizer with an initial learning rate of $4\times10^{-5}$, a weight decay of 0.05, a learning rate scheduler that uses linear learning rate decay and a linear warm-up of 1,500 iterations. Models are trained with batch size of 16 for 160K iterations. We follow the mmsegmentation codebase to adopt augmentations of random horizontal flipping, random re-scaling within ratio range [0.5, 2.0] and a random photometric distortion. Stochastic depth with ratio of $0.3$ is applied for all models. A layer-wise learning rate decay~\cite{bao2021beit} of 0.95 is adopted for all experiments.

\section{Experimental Settings for System-Level Comparison}

\subsection{SwinV2-B and SwinV2-L Settings}

Table 2, 3 and 4 include results of SwinV2-B and SwinV2-L. For these experiments, we first conduct ImageNet-22K pre-training, and then fine-tune the pre-trained models on individual down-stream recognition tasks.

\paragraph{ImageNet-22K pre-training} 

Both models use an input image size (window size) of 192$\times$192 (12$\times$12). 
We employ an AdamW optimizer~\cite{loshchilov2017decoupled} for 90 epochs using a cosine learning rate scheduler with 5-epoch linear warm-up. A batch size of 4096, an initial learning rate of 0.001, a weight decay of 0.1, and gradient clipping with a max norm of 5.0 are used. Augmentation and regularization strategies include RandAugment~\cite{cubuk2020randaugment}, Mixup~\cite{zhang2017mixup}, Cutmix~\cite{yun2019cutmix}, random erasing~\cite{zhong2020random} and stochastic depth~\cite{huang2016deep} with ratio of 0.2.

\paragraph{ImageNet-1K image classification} We consider input image sizes of 256$\times$256 and 384$\times$384. The training length is set 30 epochs, with a batch size of 1024, a cosine decay learning rate scheduler with an initial learning rate of $4\times10^{-5}$, and a weight decay of $1\times10^{-8}$. The ImageNet-1K classification weights are also initialized from the corresponding ones in the ImageNet-22K model.

\paragraph{COCO object detection}
We adopt HTC++~\cite{chen2019htc,liu2021swin} for experiments. In data pre-processing, Instaboost~\cite{fang2019instaboost}, a multi-scale training~\cite{ghiasi2019fpn} with an input image size of 1536$\times$1536, a window size of 32$\times$32, and a random scale between $[0.1, 2.0]$ are used. An AdamW optimizer~\cite{loshchilov2017decoupled} with an initial learning rate of $4\times10^{-4}$ on batch size of 64, a weight decay of 0.05, and a $3\times$ scheduler are used. The backbone learning rate is set $0.1\times$ of the head learning rate. In inference, soft-NMS~\cite{Bodla2017softnms} is used. Both single-scale and multi-scale test results are reported.

\paragraph{ADE20K semantic segmentation}
The input image size (window size) is set 640$\times$640 (40$\times$40). We employ an AdamW~\cite{loshchilov2017decoupled} optimizer with an initial learning rate of $6\times10^{-5}$, a weight decay of 0.05, a linear decayed learning rate scheduler with 375-iteration linear warm-up. The model is trained with batch size of 64 for 40K iterations. We follow the default settings in mmsegmentation for data augmentation, including random horizontal flipping, random re-scaling within ratio range $[0.5, 2.0]$ and random photometric distortion. Stochastic depth with ratio of $0.3$ is applied. 

\subsection{SwinV2-G Settings}

\paragraph{Stage-1 self-supervised pre-training}
The model is first pre-trained using a self-supervised learning approach~\cite{anonymous} on the ImageNet-22K-ext dataset (70 million images) for 20 epochs. To reduce experimental overheads, we adopt a smaller image size of 192$\times$192. The model is trained using the AdamW~\cite{loshchilov2017decoupled} optimizer with a cosine decay learning rate scheduler with 30000 steps of linear warm-up. A batch size of 9216, an initial learning rate of $1.4\times10^{-3}$, a weight decay of 0.1, and gradient clipping with a max norm of 100.0 are used. A light data augmentation strategy is employed: random resize cropping with scale range of [0.67, 1] and a aspect ratio range of [3/4, 4/3], followed by a random flipping and a color normalization steps.

\paragraph{Stage-2 supervised pre-training}
The model is further pre-trained using the class labels on the ImageNet-22K-ext dataset. We employ an AdamW~\cite{loshchilov2017decoupled} optimizer for 30 epochs, using a cosine decayed learning rate scheduler with 20000 steps of linear warm-up. A batch size of 9216, an initial learning rate of $1.4\times10^{-3}$, a layer-wise learning rate decay of 0.87, a weight decay of 0.1, and gradient clipping with a max norm of 100.0 are used. Augmentation and regularization strategies include RandAugment~\cite{cubuk2020randaugment}, random erasing~\cite{zhong2020random} and a stochastic depth~\cite{huang2016deep} ratio of 0.3.

\paragraph{Fine-tuning on ImageNet-1K image classification}
We adopt an input image size of 640$\times$640 for experiments. An AdamW~\cite{loshchilov2017decoupled} optimizer is employed for 10 epochs, using a cosine decayed learning rate scheduler and a 2-epoch linear warm-up. A batch size of 576, an initial learning rate of $2.1\times10^{-5}$, a weight decay of 0.1, and gradient clipping with a max norm of 100.0 are used. Augmentation and regularization strategies include RandAugment~\cite{cubuk2020randaugment}, random erasing~\cite{zhong2020random} and a stochastic depth~\cite{huang2016deep} ratio of 0.5.

In evaluation, we test top-1 accuracy on both ImageNet-1K V1 and V2.

\paragraph{Fine-tuning on COCO object detection}
We first conduct inter-mediate fine-tuning using the Objects-365 V2 dataset. In this stage, we remove the mask branch of the HTC++ framework~\cite{chen2019htc,liu2021swin} because there are no mask annotations. The input image resolution and window size are set as $[800, 1024]$ and $32\times 32$, respectively. In training, an  AdamW~\cite{loshchilov2017decoupled} optimizer with initial learning rate of $1.2\times10^{-3}$, a weight decay of 0.05 and a batch size of 96 are used, and the training length is set 67,500 steps.

Then we fine-tune the HTC++ model on COCO dataset, with the mask branch randomly initialized and other model weights loaded from the Objects-365-V2 pre-trained model. In this training stage, the input image resolution is set 1536$\times$1536 with a multi-scale ratio of $[0.1, 2.0]$. The window size is set 32$\times$32. The  AdamW~\cite{loshchilov2017decoupled} optimizer is employed, with an initial learning rate of $6\times10^{-4}$, a weight decay of 0.05, and a batch size of 96, and is trained 45,000 steps.

In test, Soft-NMS~\cite{Bodla2017softnms} is used. Both window sizes of $32\times32$ and $48\times 48$ are considered.
 
\paragraph{Fine-tuning on ADE20K semantic segmentation}
The input image size (window size) is set 640$\times$640 (40$\times$40). An AdamW optimizer~\cite{loshchilov2017decoupled} is employed, with an initial learning rate of $4\times10^{-5}$, a weight decay of 0.05, a linear decayed learning rate scheduler with 80K iterations, a batch size of 32, and a linear warm-up of 750 iterations. For augmentations, we follow the default settings in mmsegmentation to include random horizontal flipping, random re-scaling within ratio range $[0.5, 2.0]$ and random photometric distortion. The stochastic depth ratio is set $0.4$.

\paragraph{Fine-tuning on Kinetics-400 video action recognition}
A 2-stage fine-tuning process is employed. In the first stage, an input resolution of 256$\times$256$\times$8 with 16$\times$16$\times$8 window size is adopted. We employ the AdamW optimizer for 20 epochs using a cosine decayed learning rate scheduler with 2.5-epoch linear warm-up. Other training hyper-parameters are: batch-size 80, an initial learning rate of $3.6\times10^{-4}$, and a weight decay of 0.1.

In the second stage, we further fine-tune the model using a larger input video resolution of 320$\times$320$\times$8 with 20$\times$20$\times$8 window size. We employ the AdamW optimizer for 5 epochs using a cosine decayed learning rate scheduler with 1-epoch linear warm-up. A batch-size of 64, an initial learning rate of $5\times10^{-5}$ and a weight decay of 0.1 are set.

\section{Learnt Relative Position Bias by Different Approaches}

\begin{figure*}[ht]
    \centering
    \includegraphics[width=1.0\linewidth]{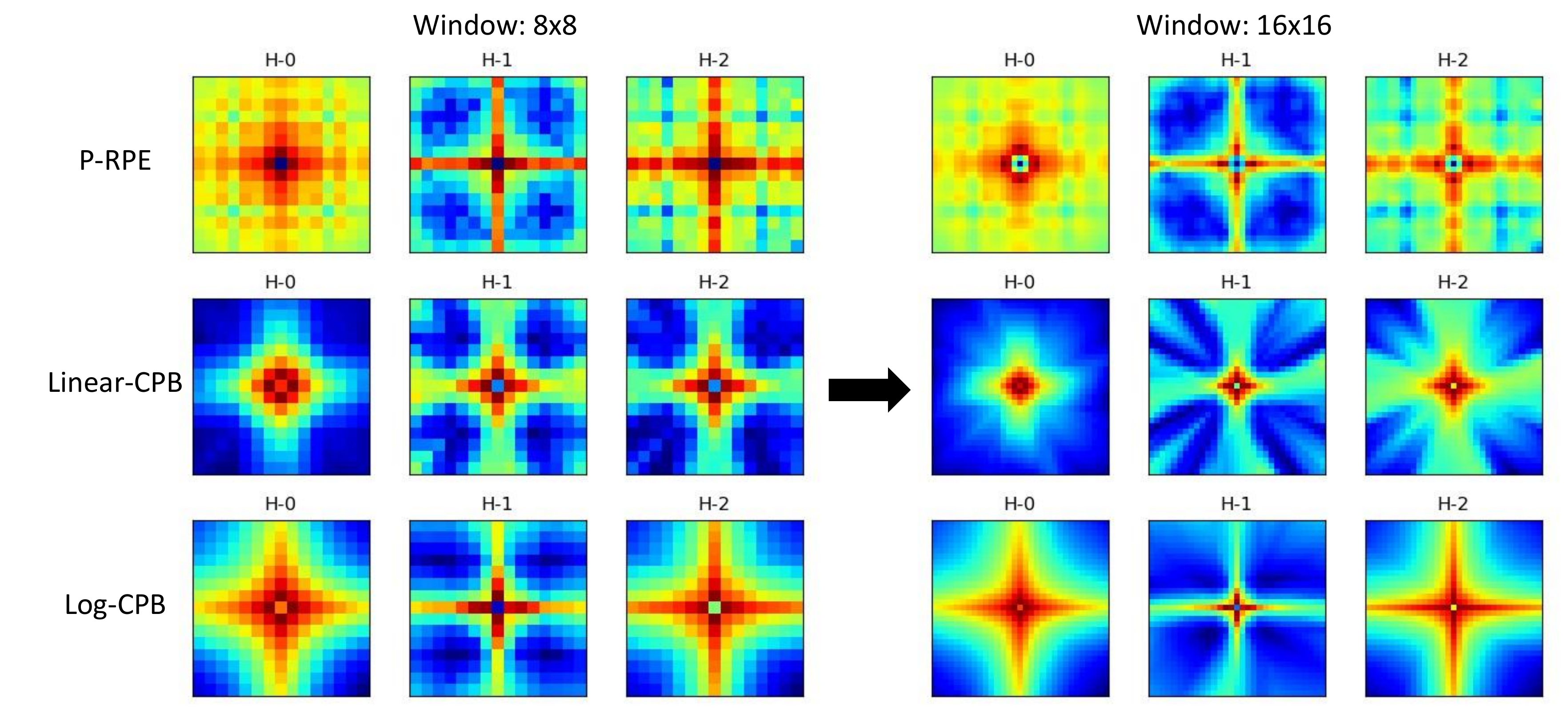}
    \caption{Visualization of the learnt relative position bias matrices by different approaches, using a SwinV2-T model and the 3 heads in the first block. Left: the bias matrices by pre-training on a 256$\times$256 image and a 8$\times$8 window; Right: the bias matrices after fine-tuning using a 512$\times$512 image size and 16$\times$16 window size. H-x indicates the x-th head.}
    \label{fig:rpe_s0b0}
\end{figure*}

\begin{figure*}[ht]
    \centering
    \includegraphics[width=1.0\linewidth]{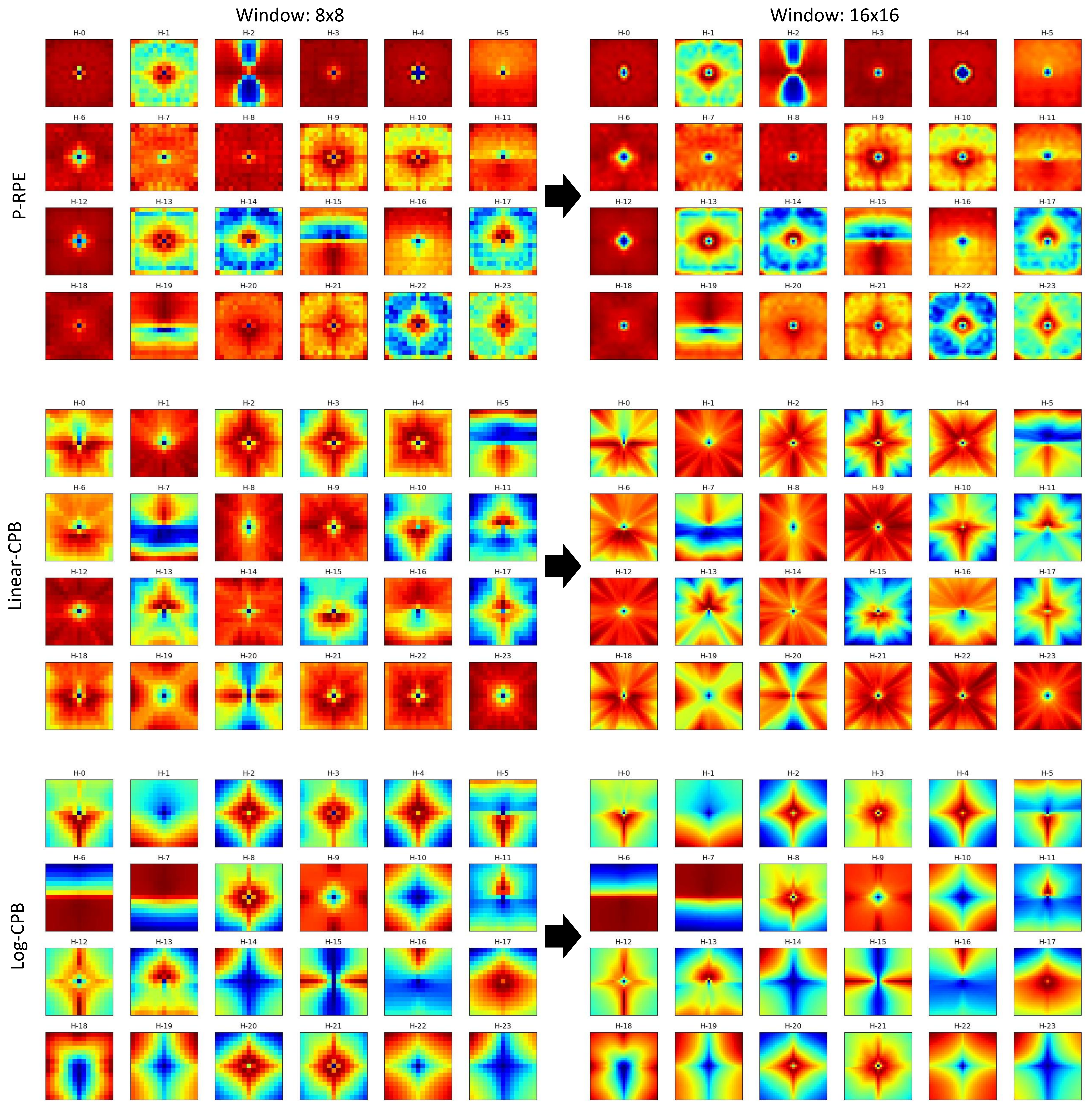}
    \caption{Visualization of the learnt relative position bias matrices by different approaches, using a SwinV2-T model and the 24 heads in the last block. Left: the bias matrices by pre-training on a 256$\times$256 image and a 8$\times$8 window; Right: the bias matrices after fine-tuning using a 512$\times$512 image size and 16$\times$16 window size. H-x indicates the x-th head.}
    \label{fig:rpe_s3b0}
\end{figure*}

Figure~\ref{fig:rpe_s0b0} visualizes the relative position bias matrices ($\hat{B} \in \mathbb{R}^{(2M-1)\times (2M-1)}$) learnt by different bias computation approaches, using a SwinV2-T model. The bias matrices of the 3 heads in the first block are visualized. The left shows the bias matrices learnt by using an input image size of 256$\times$256 and a window size of $8\times 8$. The right shows the bias matrices after fine-tuning on a larger input image resolution of 512$\times$512 and a larger window size of 16$\times$16. It turns out that the bias matrices learnt by two CPB(continuous position bias) approaches are more smoothly than that learnt by P-RPE (parameterized relative position bias). Figure~\ref{fig:rpe_s3b0} shows more examples using the last block of this model.

{\small
\bibliographystyle{ieee_fullname}
\bibliography{egbib}
}

\end{document}